\documentclass[11pt]{article}

\usepackage[preprint]{acl}

\usepackage{times}
\usepackage{latexsym}
\usepackage{makecell}
\usepackage[T1]{fontenc}

\usepackage[utf8]{inputenc}

\usepackage{microtype}

\usepackage{inconsolata}

\usepackage{graphicx}

\usepackage{hyperref}       
\usepackage{url}            
\usepackage{booktabs}       
\usepackage{caption}      
\usepackage{amsfonts}       
\usepackage{subcaption}     
\usepackage{amsthm}         
\newtheorem{definition}{Definition}
\newtheorem{remark}{Remark}
\usepackage{bbm} 
\usepackage{subcaption}     
\usepackage{multirow}       
\usepackage{graphicx}       
\usepackage{amssymb}        
\usepackage{amsmath}        
\usepackage{nicefrac}       
\usepackage{microtype}      
\usepackage{xcolor}         

\title{Shorter but not Worse: Frugal Reasoning via Easy Samples as Length Regularizers in Math RLVR}

\author{
    \textbf{Abdelaziz Bounhar}$^{1}$$^\dagger$,
    \textbf{Hadi Abdine}$^{1}$,
    \textbf{Evan Dufraisse}$^{1}$,
    \textbf{Ahmad Chamma}$^{1}$,\\
    \textbf{Amr Mohamed}$^{1}$,
    \textbf{Dani Bouch}$^{1}$,
    \textbf{Michalis Vazirgiannis}$^{1,2}$,
    \textbf{Guokan Shang}$^{1}$$^\dagger$
    \\
    \\
    $^{1}$MBZUAI, $^{2}$Ecole Polytechnique
    \\
    \small{
        $^\dagger$Correspondence: \texttt{\{abdelaziz.bounhar, guokan.shang\}@mbzuai.ac.ae}
    }
}

\begin{document}

\maketitle
\begin{abstract}
Large language models trained for step-by-step reasoning often become excessively verbose, raising inference cost. 
Standard Reinforcement Learning with Verifiable Rewards (RLVR) pipelines conventionally filter out ``easy'' problems for training efficiency, leaving the model to train primarily on harder problems that require longer reasoning chains. 
This skews the output length distribution upward, resulting in a model that conflates ``thinking longer'' with ``thinking better''. 
We show that retaining and modestly up-weighting moderately easy problems acts as an implicit length regularizer: rewards become associated with concise solutions early, preventing runaway verbosity without any explicit length penalization. 
RLVR experiments using this approach on \texttt{Qwen3-4B-Thinking-2507} and \texttt{Qwen3-30B-A3B-Thinking-2507} achieve baseline pass@1 accuracy on AIME25 benchmark while generating solutions that are, on average, nearly twice as short. 
The two resulting \texttt{\textbf{Frugal-Thinking}} models and all associated code and data are publicly available\footnote{\url{https://hf.co/collections/MBZUAI-Paris/frugal-thinking}}.
\end{abstract}

\section{Introduction}

Recently, Large Language Models (LLMs) have begun to rapidly advance the frontier of machine intelligence through test time scaling via step-by-step ``reasoning''. 
Scaling inference budget and training with RLVR have enabled models to achieve strong performance on competition-level mathematics and coding tasks by producing extended chains of thought. 
However, this progress often incurs at a cost: reasoning models tend to be overly verbose, generating excessively long solutions that increase inference latency and memory usage.  

A common design choice in RLVR training pipelines is to filter out ``easy'' problems to maximize training efficiency, with training typically beginning in medium-difficulty samples and gradually shifting toward harder instances \citep{magistral, amthinkingv1_advancingfrontierreasoning,am_team_difficultyawarestagedreinforcementlearning}. 
This design choice is not arbitrary; It follows from the mechanics of Group Relative Policy Optimization (GRPO) \citep{grpo}, wherein groups with either all-correct or all-incorrect rollouts yield zero advantage and therefore provide no learning signal.
Consequently, both easy and unsolvable hard problems are typically excluded, as they are unlikely to produce meaningful policy updates.
This leaves the model to learn primarily from problems that inherently require longer reasoning chains. 
Over time, this imbalance shifts the output length distribution upward, leading the policy to reward verbosity even when many of the generated tokens are redundant.
The outcome is a systematic drift towards unnecessarily long outputs, where models conflate ``\textit{thinking longer}'' with ``\textit{thinking better}''.  
A complementary information-theoretic intuition is that longer prefixes can reduce uncertainty about the final answer when length is unpenalized (formalized in Section~\ref{sec:prelude}), which can inadvertently reward verbosity.

In this paper, we revisit the training-efficiency heuristic of discarding easy problems and instead argue for their value. 
We show that retaining, and upweighting moderately easy problems provides a natural counterbalance: they act as length regularizers.
By exposing the policy to tasks that admit concise solutions and training under a limited context window, the model is implicitly pressured to maintain efficiency in its output distribution in order to obtain rewards on harder samples.
This encourages the policy to preserve correctness while keeping solutions compact under a fixed context budget.
Experiments on \texttt{Qwen3-4B-Thinking-2507}\footnote{\url{https://hf.co/Qwen/Qwen3-4B-Thinking-2507}} with a 16k-token budget show that our method preserves baseline pass@1 accuracy on AIME25, while reducing solution length by nearly (2$\times$).
This demonstrates that concision and performance are not in opposition: carefully curating the training data is and has always been the key.

This work has two primary contributions:
\begin{itemize}
    \item \textbf{Implicit length regularization:}
    We show that emphasizing moderately easy problems in RLVR training naturally regularizes output length, reducing verbosity without explicit reward shaping.
    \item \textbf{Empirical validation:} 
    With a 16k-token budget, and on the dense model (\texttt{Qwen3- 4B-Thinking-2507}), our method preserves baseline pass@1 accuracy on AIME25 while nearly halving the average output solution length. 
    On the Mixture of Experts (MoE) model (\texttt{Qwen3-30B-A3B-Thinking-2507}\footnote{\url{https://hf.co/Qwen/Qwen3-30B-A3B-Thinking-2507}}), performance remains largely preserved, with better token efficiency and while delivering consistent improvements across other benchmarks.
\end{itemize}

Together, these findings highlight that data curation, not only reward design or model size, plays a critical role in shaping the efficiency of reasoning language models. 

\section{Prelude}
\label{sec:prelude}
We consider an autoregressive language model parameterized by $\theta$, defining a policy
$\pi_\theta$ over token sequences. For a query $x$ and a response $y = (y_1,\dots,y_T)$, the likelihood under the policy is
\begin{equation}
\pi_\theta(y|x) = \prod_{t=1}^T \pi_\theta(y_t \mid x, y_{<t}).
\end{equation}
Each response is evaluated by a verifier $r(x, y)$ defined over appropriate domains, which assigns a scalar reward indicating correctness\footnote{Throughout this work, we use binary verifiable rewards $r(x, y) \in \{0, 1\}$, where $1$ denotes a correct solution.}.
RLVR seeks to optimize $\pi_\theta$ so as to maximize the expected verifier score $\mathbb{E}_{x \sim \mathcal{D},y \sim \pi_\theta}[r(x, y)]$ where $\mathcal{D}$ is the training dataset.

\paragraph{Group Relative Policy Optimization (GRPO).}
Instead of relying on a value model as in PPO \citep{ppo}, GRPO \citep{grpo}
uses groups of $G$ responses $\{y_i\}_{i=1}^G \sim \pi_{\theta_{\text{old}}}(\cdot|x)$ for the same query $x$ sampled from a training dataset $\mathcal{D}$ to estimate the expected reward, a.k.a. the value function.
\begin{equation}
\label{eq:grpo}
\begin{aligned}
&J_{\text{GRPO}}(\theta)
=
\\
&\hspace{0.2cm} \mathbb{E}_{x \sim \mathcal{D}, \{y_i\}_{i=1}^G \sim \pi_{\theta_{\text{old}}}(\cdot \mid x)}
\Bigg[
\frac{1}{G}
\sum_{i=1}^G
\frac{1}{|y_i|}
\sum_{t=1}^{|y_i|}
\\
&\hspace{0.2cm}\min \Big(
w_{i,t}(\theta) A_i,
\;
\text{clip}\!\left(
w_{i,t}(\theta),
1-\epsilon,
1+\epsilon
\right) A_i
\Big)
\Bigg].
\end{aligned}
\end{equation}
where each response receives an advantage computed relative to the group:
\begin{equation}
\label{eq:adv_def_grpo}
A_i = \frac{r(x,y_i) - \text{mean}\left(\{r(x,y_i)\}_{i=1}^G\right)}{\text{std}\left(\{r(x,y_i)\}_{i=1}^G\right)},
\end{equation}
and
\begin{equation}
w_{i,t}(\theta) = \frac{\pi_\theta(y_{i,t} \mid x,y_{i,<t})}{\pi_{\theta_{\text{old}}}(y_{i,t} \mid x,y_{i,<t})}
\end{equation}
is the \emph{importance sampling weight} applied at the token level. 
As in PPO, GRPO uses token-level importance ratios with clipping to stabilize updates under target-behavioral policies distribution shifts.

\paragraph{Difficulty and vanishing advantage.}
By construction, if all $G$ rollouts are correct ($r=1$ for all) or all are incorrect ($r=0$ for all), then $A_i = 0$ for every $i$, see \eqref{eq:adv_def_grpo}. 
Such groups do not provide a gradient signal. 
Consequently, ``easy'' problems (solved with probability $\approx 1$) and ``hard'' problems (solved with probability $\approx 0$) are systematically excluded from the RLVR pipelines. 
Training is therefore efficient when done on samples that satisfy
\begin{equation}
0 < \Pr \left[r(x,y)=1 \right] < 1.
\end{equation}

\paragraph{Length bias from difficulty imbalance.}
Medium and hard problems inherently require longer reasoning chains.
Filtering out easy problems
therefore biases the effective training distribution toward longer outputs.
Over successive updates, the policy may learn that reward is typically associated with extended completions, skewing the output length distribution upward. Empirically, this manifests as models producing unnecessarily long solutions, even when concise reasoning would suffice.

\paragraph{Information-theoretic view.}
Let $x \in \mathcal{X}$ be a query, $Y$ denote the (final) answer token (or a deterministic function of the full response), 
and $Z_t \triangleq (Y_1,\dots,Y_t)$ be the length-$t$ prefix produced by the autoregressive policy. 
We treat $(x,Y,Z_t)$ as jointly distributed under the rollout process. 
By the chain rule of entropy,
\begin{align}
0 &\le\; I\!\left(Y; Y_{t+1} \,\mid \, X,Z_t\right) \nonumber \\
&= \mathcal{H}(Y \mid X,Z_t) - \mathcal{H}(Y \mid X,Z_{t+1}),
\end{align}
so $\mathcal{H}(Y \mid X,Z_{t+1}) \le \mathcal{H}(Y \mid X,Z_t)$ for all $t$. 
That is, \emph{conditioning on a longer prefix reduce the conditional entropy of the final answer}. 
This statement holds irrespective of how $Y_{t+1}$ is generated (it may even be semantically vacuous), 
as it is simply a property of conditional entropy and mutual information.

\smallskip

\noindent\underline{\emph{Implication for RLVR.}} 

In the absence of any explicit penalty on output length or token semantics, and with rewards depending solely on the correctness of the final answer, since conditioning on a longer prefix can only decrease $\mathcal{H}(Y \mid X, Z_t)$, a policy can (weakly) reduce uncertainty by delaying commitment to the final answer, especially on medium/hard instances where
longer chains correlate with higher success. Over training, this can bias the learned output distribution toward longer completions.
In effect, \textit{verbosity becomes a statistical shortcut to entropy reduction rather than a reflection of genuine reasoning}.


\section{Methodology}
\label{sec:method}

This section describes our reinforcement learning setup for mathematical reasoning, including the reward definition and a two-stage data curation strategy. The design is motivated by the relationship between problem difficulty, solution length, and effective learning signal.

\paragraph{Length regularization via moderately easy problems.}
Consider a problem with a success-rate parameter $p$ defined as
\begin{equation}
p = \Pr\left[r(x,y)=1 \mid y \sim \pi_\theta \right].
\end{equation}
The easy problems correspond to $p \approx 1$, and the hard ones to $p \approx 0$.
%
We retain all problems with $p < 1$, so that the policy receives frequent positive signal from concise, solvable trajectories, which regularizes the output length distribution under a fixed context window.

\paragraph{Reward shaping.}
We adopt a binary verifiable reward based on exact string matching of the extracted final answer. 
The model encloses its final prediction within \texttt{\textbackslash boxed\{\}}, allowing deterministic parsing and verification.
Let $\hat{a}(x)$ denote the model’s predicted answer extracted from its output for a given query $x$.
A reward of $1$ is assigned if the normalized prediction matches the normalized ground truth $y$, and $0$ otherwise:
\begin{equation}
\label{eq:reward_fct}
r(x,y) =
\begin{cases}
1, & \text{if } \hat{a}(x) = y, \\[6pt]
0, & \text{otherwise.}
\end{cases}
\end{equation}

\begin{figure*}[t]
    \centering
    \includegraphics[width=0.49\textwidth]{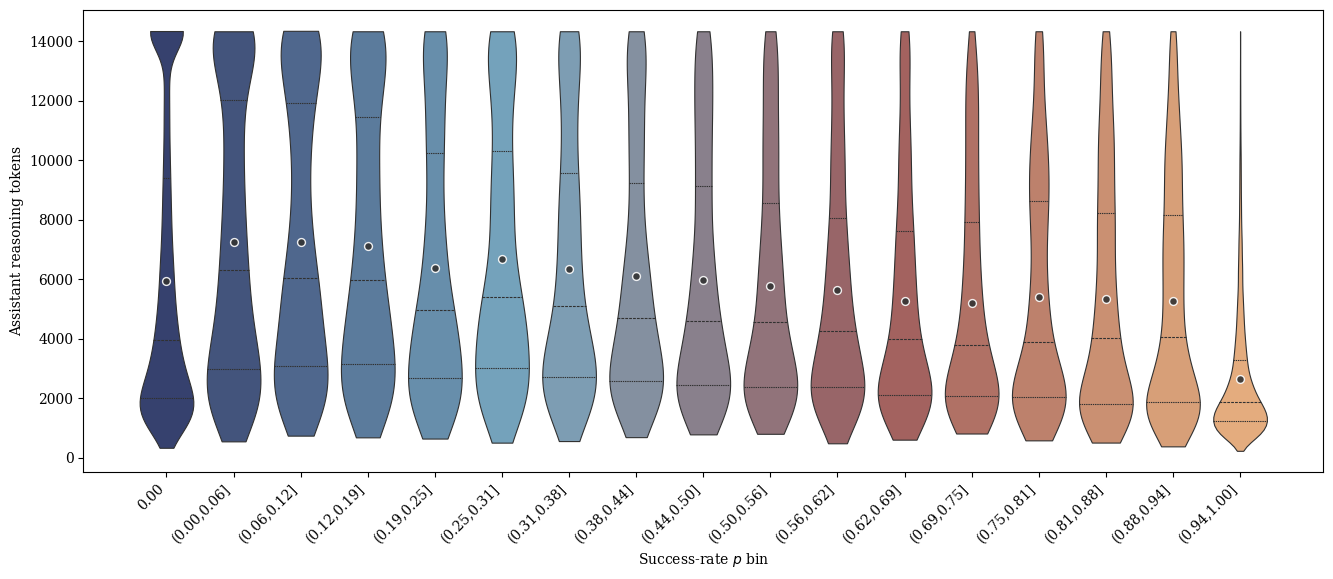}
    \hfill
    \includegraphics[width=0.48\textwidth]{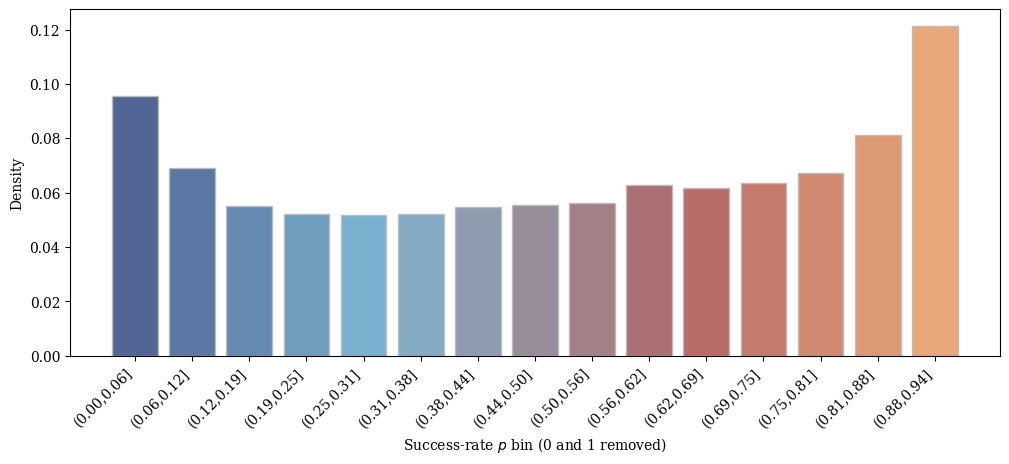}
    \caption{
    Empirical success-rate analysis.
    \textbf{Left:} Token count distribution as a function of empirical success rate \(p\).
    \textbf{Right:} Difficulty distribution \(\rho(p)\) after filtering out trivial and unsolvable cases (\(p \in \{0,1\}\)).
    }
    \label{fig:success_rate_analysis}
\end{figure*}


\paragraph{Data curation stage 1: emergent brevity.}
We started with the collection of maths data curated by \citep{amthinkingv1_advancingfrontierreasoning, am_team_difficultyawarestagedreinforcementlearning}.
Although batches are sampled uniformly at random, with mixed difficulty\footnote{Curriculum learning is applied only in the second stage.}, the global dataset distribution is deliberately imbalanced, containing a higher proportion of moderately easy problems.
%
%
Because short, solvable problems provide stable positive rewards associated with structured and concise reasoning traces, they tend to dominate the effective reward signal.
Very hard problems, by contrast, contribute little due to sequence truncation or verification failure when the policy cannot yet solve them.
Over time, this inductive bias in the reward signal implicitly encourages shorter, more efficient reasoning traces.  

Formally, the expected reward over the training distribution $\mathcal{D}$ can be expressed as
\begin{equation}
\mathbb{E}_{(x,y)\sim\mathcal{D}}[r(x,y)]
= \int_0^1 p \ \rho(p) \ dp,
\end{equation}
where $\rho(p)$ denotes the empirical density of problems with success probability $p$.
In our dataset, $\rho(p)$ is intentionally skewed toward easy problems, so these samples dominate the reward signal.
During RL, the gradient of $J_{\text{GRPO}}(\theta)$ therefore receives stronger, more stable updates from solvable problems within the token limit, while hard problems often yielding long or truncated completions contribute negligible gradients.
This imbalance constrains the learned output distribution: since rewards arise predominantly from shorter, solvable trajectories, verbosity ceases to be a profitable strategy.
%
In short, associating reward with short, solvable trajectories acts as an implicit length regularizer that encourages concise reasoning.
The empirical success-rate distribution $\rho(p)$ 
computed using our target model \texttt{Qwen3-4B-Thinking-2507} based on 16 rollouts per-prompt exhibited a bimodal pattern, with a large mass at $p=0$ and $p=1$.
This pattern indicates that many problems are either trivially solved or currently unsolved by the base policy (given a budget of 16k tokens), while relatively few lie in the intermediate difficulty range where learning gradients are most informative.
The right side of Figure~\ref{fig:success_rate_analysis} isolates that central region by excluding samples with $p\!\in\!\{0,1\}$,
highlighting the subset that drives effective RLVR optimization.
Additionally, the left side of Figure~\ref{fig:success_rate_analysis} shows that reasoning length varies systematically with difficulty, with easy problems requiring a small number of tokens.

\begin{figure}[ht]
\centering
\includegraphics[width=0.99\columnwidth]{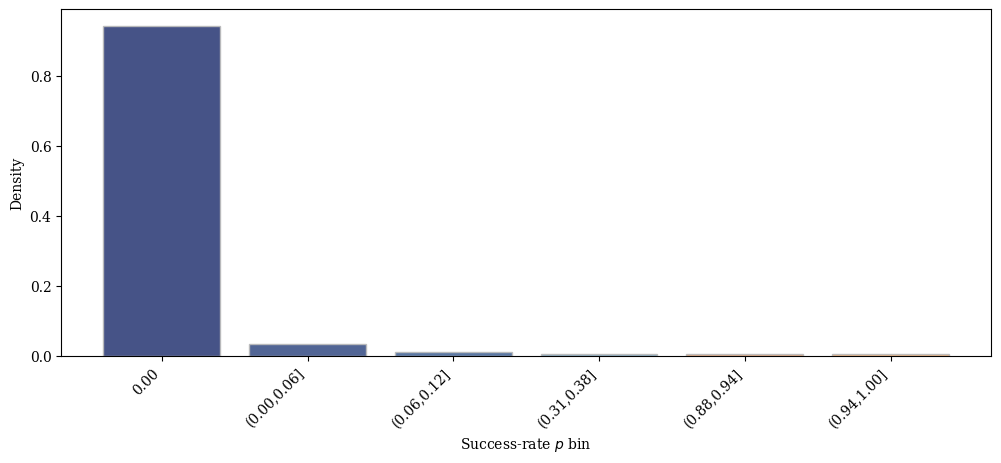}
\caption{Distribution \(\rho(p)\) after scaling maximum response length to 42k tokens for a random sample of 256 problems with $p=0$.}
\label{fig:long_context_success}
\end{figure}

\begin{remark}
Some instances that are initially unsolved at 16k become solvable with longer budgets, so we retain $p=0$ samples in Stage~1 rather than filtering them out, see Figure~\ref{fig:long_context_success}.
\end{remark}

\paragraph{Data curation stage 2: improvement via curriculum RLVR.}
Having obtained a concise and efficient policy after \textit{Stage 1}, we proceed with a second reinforcement phase based on curriculum RLVR.
Our goal in this stage is to enrich the model’s knowledge and reasoning capabilities on a wider domain of mathematical problems while maintaining the same 16k-token limit.
Training is conducted on a filtered subset of the \texttt{DeepMath-103} dataset \citep{deepmath},
which contains problems grouped by difficulty.
We follow the difficulty annotations provided by the authors to structure a progressive curriculum: training begins with moderately solvable instances and gradually incorporates harder problems as the policy’s competence improves.

We filtered the \texttt{DeepMath-103} dataset through a multi-step process. First, we removed samples already present in the Stage~1 dataset. We then retained only those with the correct format and sufficient difficulty.

Following the pre-filtering procedure from previous work\footnote{\url{https://hf.co/datasets/ChuGyouk/DeepMath-Filtered-59.9K}}, we excluded Multiple-Choice, True/False, Yes/No, and formal Proof-based questions, as in the work of \citet{magistral}. We also discarded examples with inconsistent answers across the three DeepSeek-R1 generations, those asking for counterexamples or lacking a single correct answer, and those that were ill-posed or underspecified. We adopted the annotations from the same repository, produced using \texttt{gpt-5-mini-2025-08-07} with \texttt{verbosity = "medium"} and \texttt{reasoning\_effort = "minimal"}. After this stage, the dataset was reduced from 103k to 57k samples.

Next, we filtered by difficulty. For each of the nine second-level math domains in the DeepMath dataset, we sampled around 30 examples for difficulty levels 5 to 9 and evaluated the pass@1 performance across these levels. Model accuracy varied across domains: since training datasets typically overrepresent precalculus, calculus, and algebra problems, performance was higher in those areas. For each domain, we retained all difficulty levels starting from the one with less than 75\% success-rate, yielding a final set of 14.5k samples (see Table~\ref{tab:difficulty} for cut details).
\begin{table}[ht]
\centering
\resizebox{0.99\columnwidth}{!}{%
\begin{tabular}{l c}
\toprule
\textbf{Math Domain} & \textbf{Starting Difficulty Level} \\
\midrule
Algebra & 7 \\
Calculus & 7 \\
Precalculus & 7 \\
Discrete Mathematics & 7 \\
Number Theory & 6 \\
Geometry & 7 \\
Other & 5 \\
Applied Mathematics & 6 \\
\bottomrule
\end{tabular}
}
\caption{Stage 2 Filtering - Difficulty level (start included) for each second-level math domain.}
\label{tab:difficulty}
\end{table}

Empirically, we find that performing curriculum RLVR on this subset gives us the best performances compared to random shuffling.

\begin{figure*}[t]
\centering
\setlength{\tabcolsep}{2pt}

\begin{subfigure}[t]{0.19\linewidth}
  \centering
  \includegraphics[width=\linewidth]{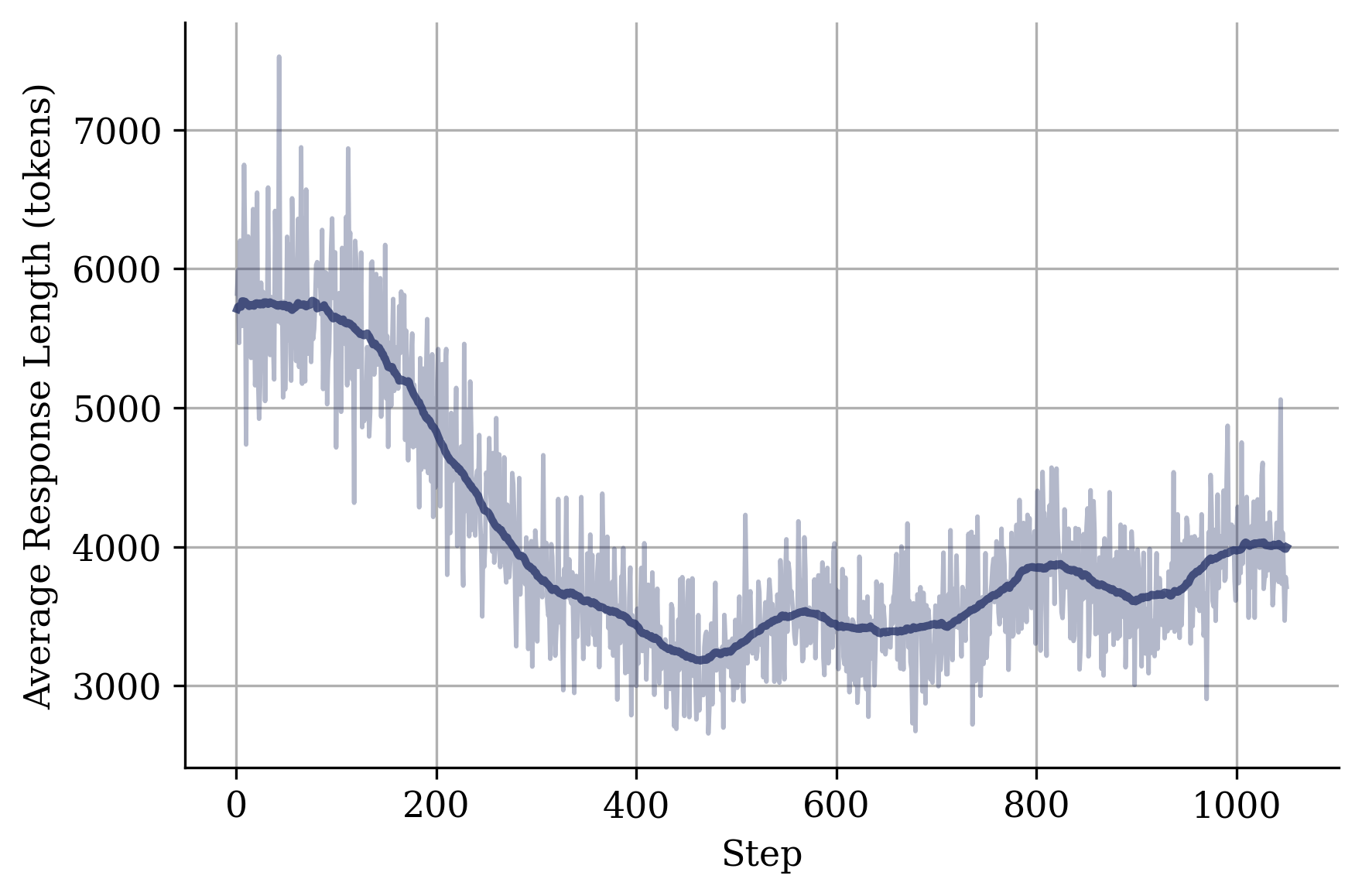}
  \caption{Avg. length.}
\end{subfigure}\hfill
\begin{subfigure}[t]{0.19\linewidth}
  \centering
  \includegraphics[width=\linewidth]{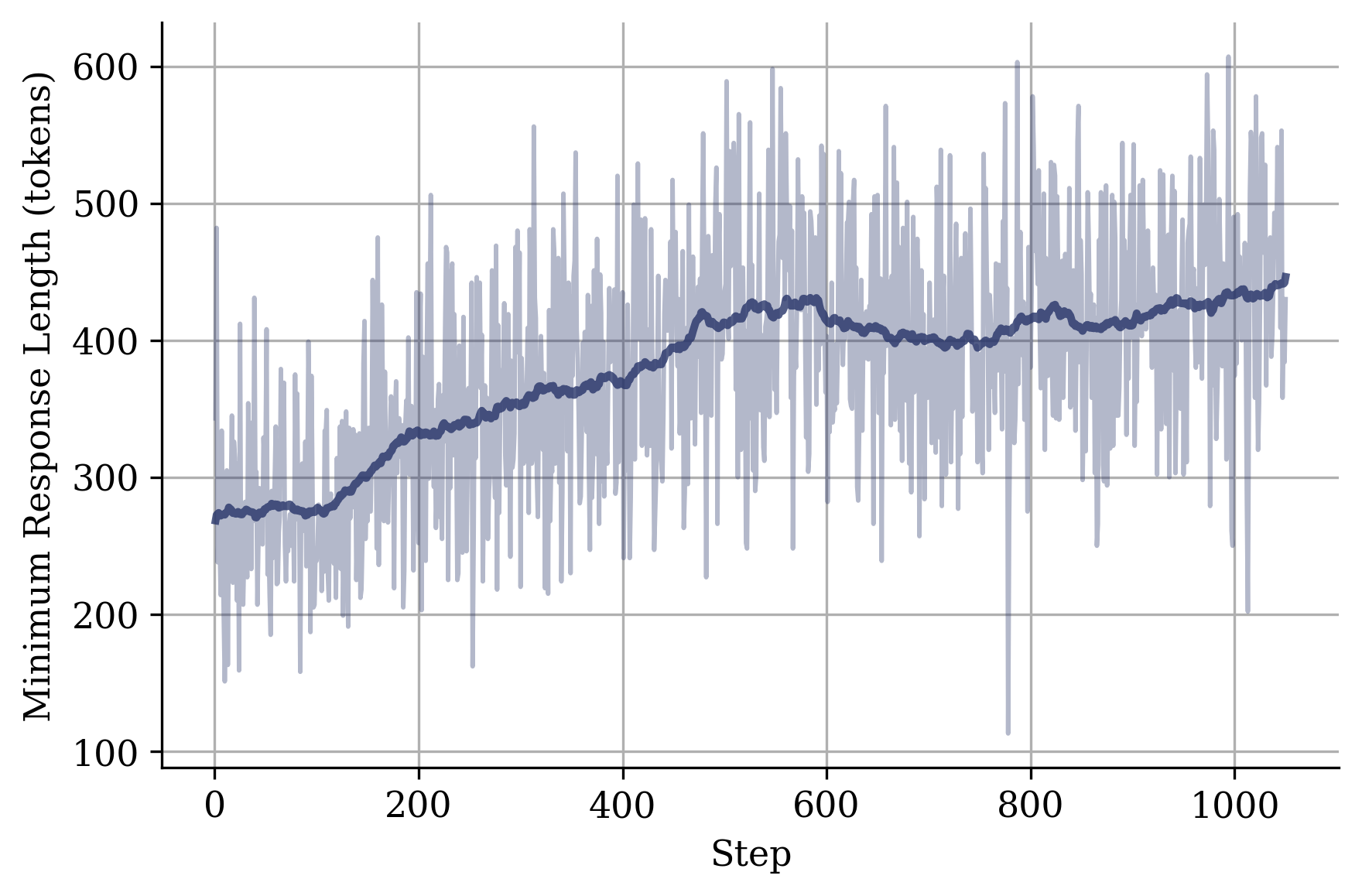}
  \caption{Min. length.}
\end{subfigure}\hfill
\begin{subfigure}[t]{0.19\linewidth}
  \centering
  \includegraphics[width=\linewidth]{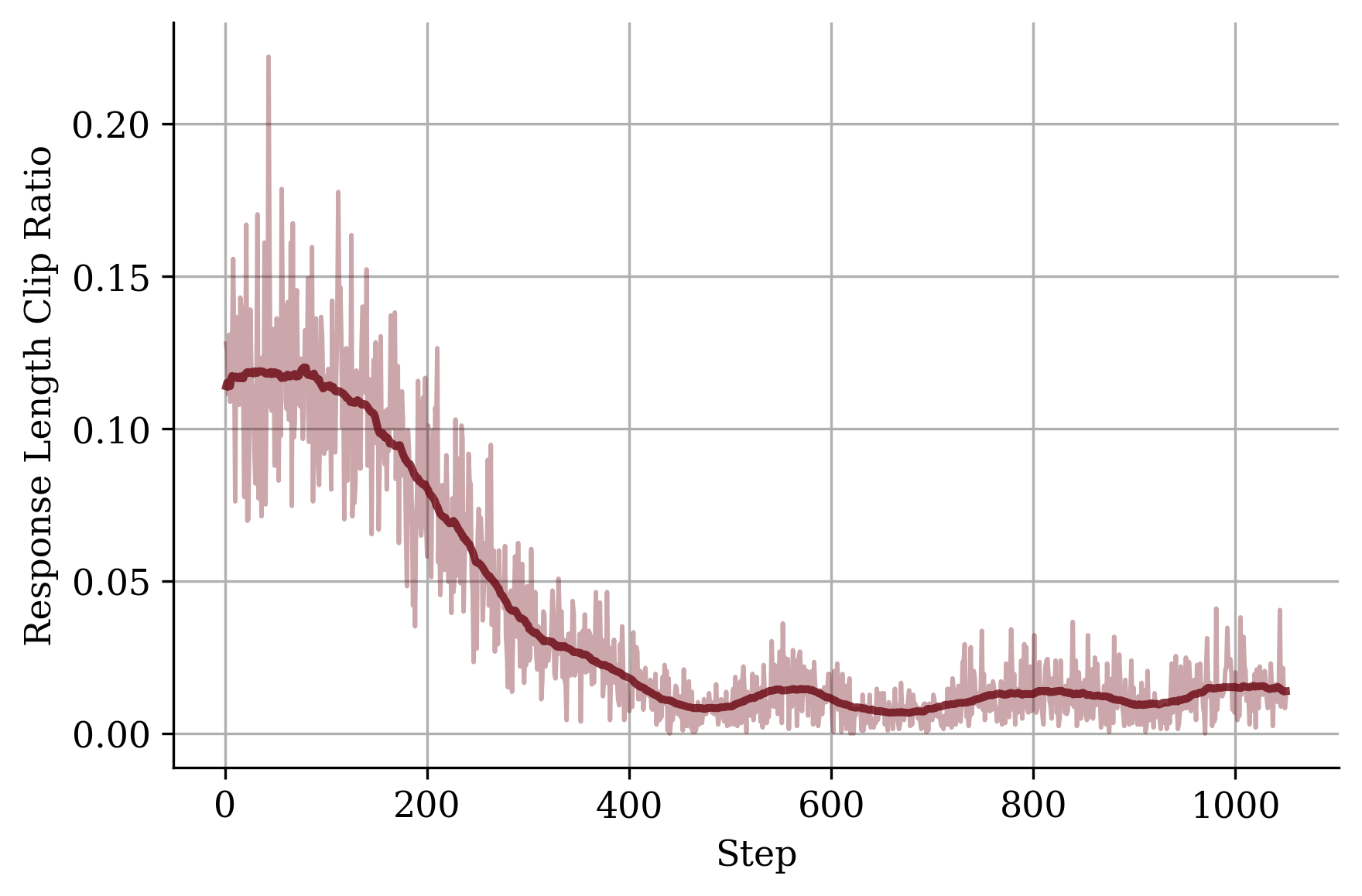}
  \caption{Clip ratio.}
\end{subfigure}\hfill
\begin{subfigure}[t]{0.19\linewidth}
  \centering
  \includegraphics[width=\linewidth]{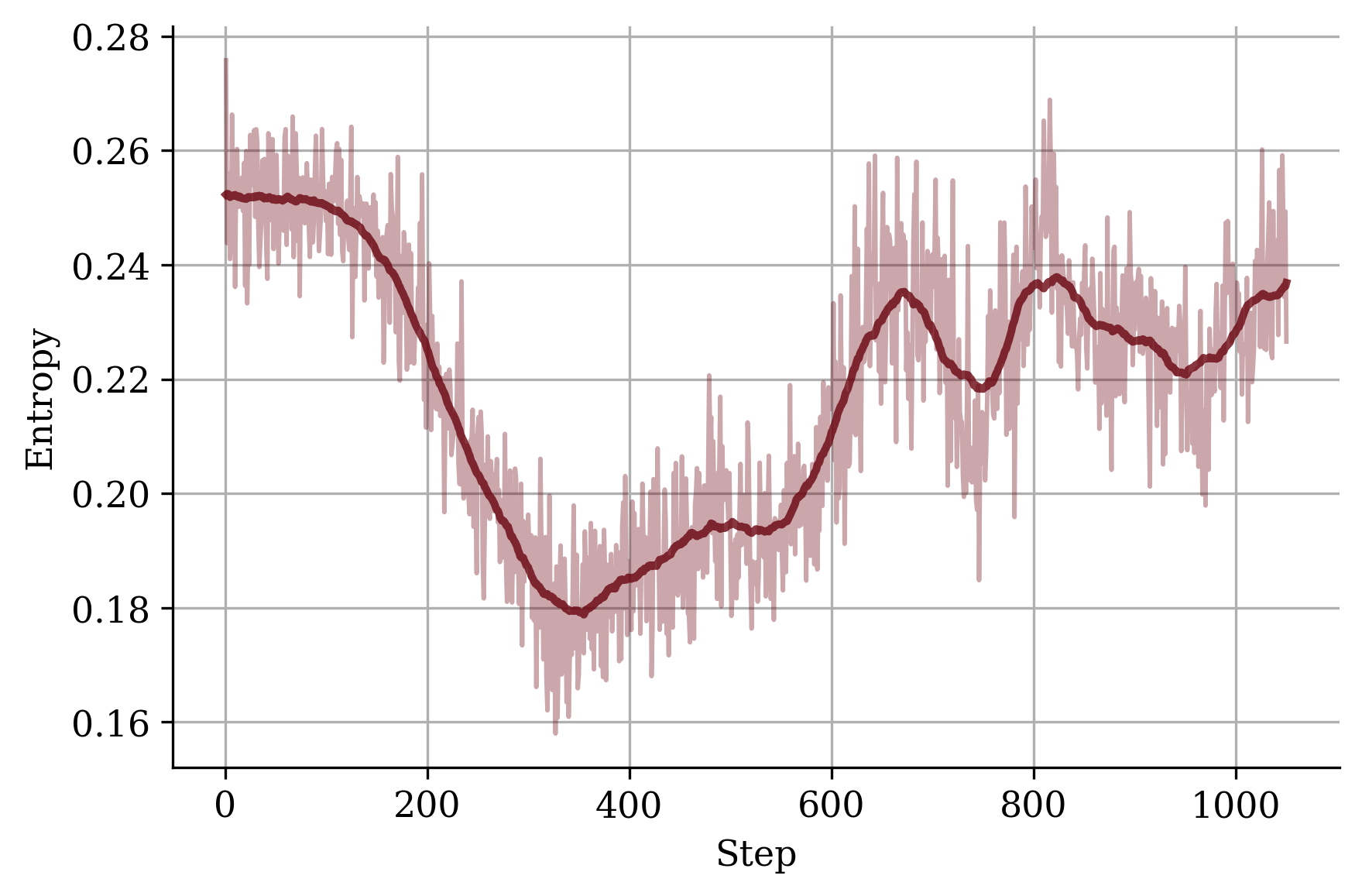}
  \caption{Entropy.}
\end{subfigure}\hfill
\begin{subfigure}[t]{0.19\linewidth}
  \centering
  \includegraphics[width=\linewidth]{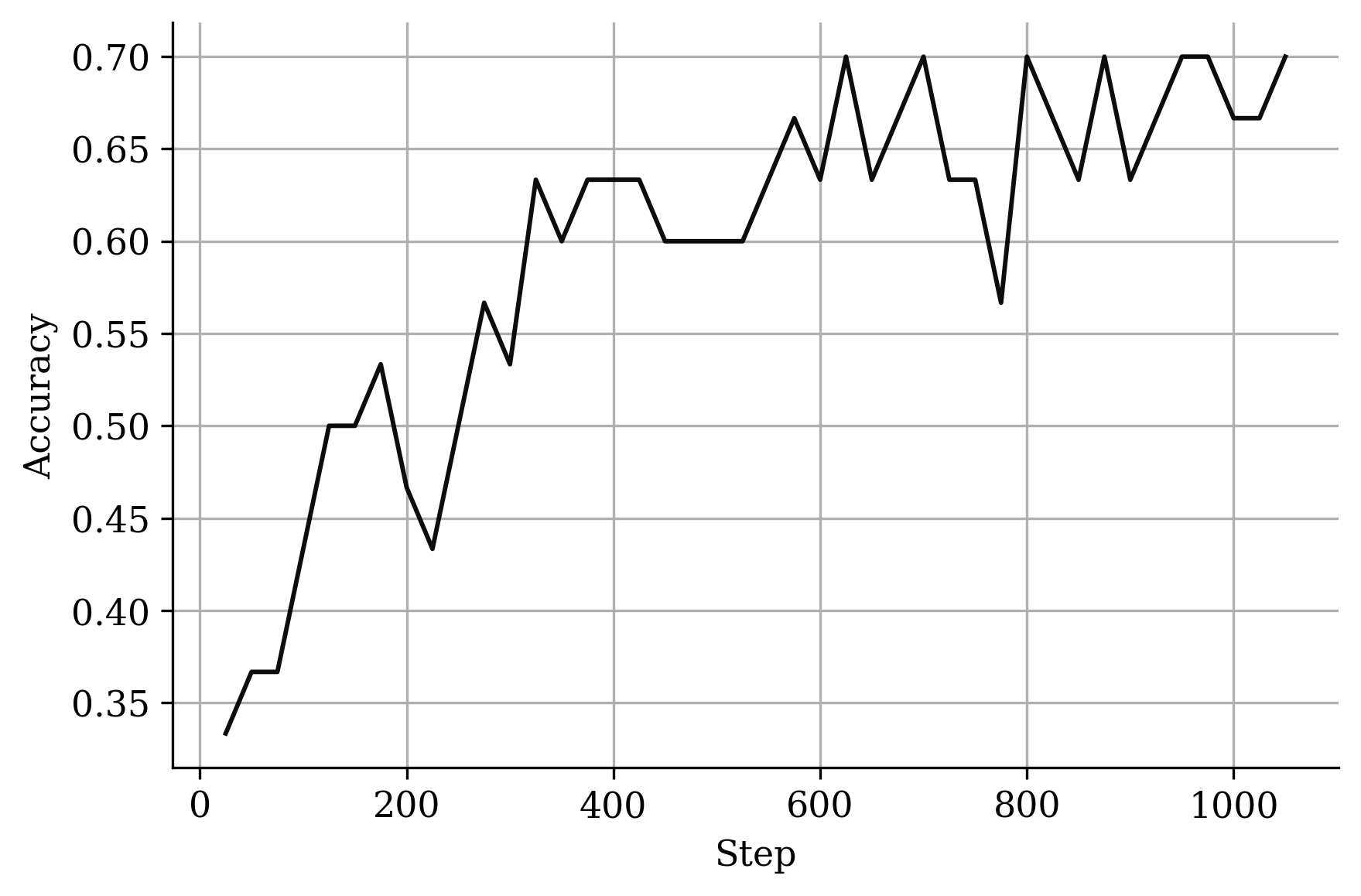}
  \caption{AIME25 acc.}
\end{subfigure}

\vspace{-0.4em}
\caption{Training dynamics during Stage~1 (emergent brevity). Early training is dominated by overly long, truncated generations with high entropy and low accuracy. As learning progresses, average response length and clip ratio decrease sharply, entropy stabilizes, and validation accuracy on AIME25 improves steadily—showing that conciseness and correctness co-emerge.}
\vspace{-0.6em}
\label{fig:training_dynamics}
\end{figure*}





\section{Experiments}

\subsection{Experimental Setup}

We fine-tune \texttt{Qwen3-4B-Thinking-2507} and \texttt{Qwen3-30B-A3B-Thinking-2507} using GRPO with a verifiable binary reward function as defined in Eq. \eqref{eq:reward_fct}. 
Our RLVR implementation is based on the \texttt{verl} framework~\citep{verl}.

Training uses the curated math datasets described in Section~\ref{sec:method} and proceeds in two stages:
\begin{itemize}
    \item \textbf{Stage~1 (emergent brevity).} 1{,}050 optimization steps (one epoch) on the Stage~1 curated dataset.
    \item \textbf{Stage~2 (curriculum RLVR).} Additional training on our filtered subset of \texttt{DeepMath-103}: 255 steps (two epochs) for the 4B model, and one epoch for the 30B MoE model, which converges with fewer updates.
\end{itemize}

All stages use a fixed 16k-token generation limit and share identical optimization hyperparameters\footnote{Stage~2 also includes a short warm-up phase before resuming full-rate training.}.
A summary of the main hyperparameters is provided in Table~\ref{tab:hyperparams}.
\begin{table}[ht]
\centering
\renewcommand{\arraystretch}{1.17} 
\resizebox{0.95\columnwidth}{!}{%
\begin{tabular}{l c}
\toprule
\textbf{Parameter} & \textbf{Value} \\
\midrule
Base model & \texttt{Qwen3-4B-Thinking-2507} \\
RL algorithm & GRPO \\
Reward type & Verifiable binary reward (exact match) \\
Rollout group size ($G$) & 16 \\
Clipping thresholds $(1-\epsilon, 1+\epsilon)$ & (0.8, 1.28) \\
Maximum completion length & 16{,}384 tokens \\
Batch size (per step) & 128 \\
Learning rate & $1 \times 10^{-6}$ \\
Warmup schedule & Linear, first 5\% of steps \\
Optimizer & AdamW \\
Hardware & 250 NVIDIA H200 GPU days \\
\bottomrule
\end{tabular}%
}
\caption{Hyperparameters and system configuration for RL fine-tuning.}
\label{tab:hyperparams}
\end{table}

\subsection{Training Dynamics}
We analyze the training dynamics of the 4B dense variant, as the 30B MoE model exhibits qualitatively similar training dynamics.
Figure~\ref{fig:training_dynamics} summarizes the evolution of key metrics during training \textit{Stage~1}.  
At the beginning of optimization, the model displays pronounced verbosity, reflected by a response-length clipping ratio exceeding 15\%, 
indicating that many generations are prematurely truncated at the 16k-token limit.
As training progresses, the average response length steadily decreases while the minimum length increases,
suggesting that the model learns to produce more compact yet complete reasoning traces.  
This reduction in verbosity coincides with a sharp decline in the response-clipping ratio, confirming that the policy increasingly completes its reasoning within the available context budget.

Entropy dynamics provide additional insight into this transition.
Entropy decreases sharply in the early phase as the policy shifts from exploration to exploitation, 
stabilizing around consistent reasoning patterns that yield reliable verifier rewards.
Around mid-training (steps 400–600), entropy rises slightly again, indicating renewed exploration which can indicate that the model begins tackling more diverse or harder samples, yet without reverting to the excessively long outputs observed initially.  
This interplay between entropy and response length supports the interpretation of \emph{emergent brevity} as a stable equilibrium: 
the policy reduces uncertainty through more efficient reasoning rather than through longer sequences.

Validation accuracy on AIME25 (right panel) increases steadily from roughly 33\%\footnote{Evaluation conducted under a 16k-token generation budget.} 
to about 70\% throughout Stage 1, showing that conciseness and reasoning competence improve in tandem rather than in opposition.
By the end of Stage 1, the policy achieves strong accuracy while maintaining concise, self-terminating outputs, 
consistent with the intended effect of implicit length regularization.  
During Stage 2 (Curriculum RLVR), the overall behavior remains qualitatively similar: 
the model continues to generate short, efficient reasoning traces.
Though, we only observed the minimum response length increasing to $\approx$ 1,200 tokens due to the increased difficulty.

\section{Evaluation}

We evaluate our method on verifiable mathematical reasoning tasks, focusing on the efficiency–accuracy trade-off induced by easy-sample regularization and curriculum RLVR.
We report both standard accuracy and our proposed Efficiency-Adjusted Accuracy (EAA; Definition~\ref{def:eaa}) to jointly assess performance and output conciseness.

\begin{table*}[ht]
\centering
\renewcommand{\arraystretch}{1.2} 
\resizebox{\textwidth}{!}{%
\begin{tabular}{l ccccccc c c}
\toprule
\textbf{Model} & 
\textbf{Size} &
\textbf{GPQA Diamond} &
\textbf{AIME25} &
\textbf{Omni-Hard} &
\textbf{GSM\_PLUS} &
\textbf{IFEVAL} &
\textbf{MATH\_500} &
\phantom{a} &
\textbf{Average} \\
\midrule
Qwen3-30B-A3B-Thinking-2507 & 30B & \textbf{70.71}$\mid$43.96 & \textbf{86.67}$\mid$13.93 & 08.09$\mid$00.63 & 90.29$\mid$\textbf{90.29} & 41.35$\mid$\textbf{41.35} & \textbf{97.80}$\mid$62.73 & & 65.82$\mid$42.15 \\
Magistral-Small-2509 & 24B & 62.63$\mid$\textbf{62.63} & 80.00$\mid$20.71 & \textbf{53.18}$\mid$11.41 & 88.86$\mid$86.42 & 39.71$\mid$30.77 & 96.60$\mid$81.77 & & 70.16$\mid$48.95 \\
Magistral-Small-2507 & 24B & 57.07$\mid$02.84 & 53.33$\mid$02.66 & 34.10$\mid$03.60 & 81.29$\mid$04.05 & 41.75$\mid$06.76 & 93.20$\mid$04.64 & & 60.12$\mid$04.09 \\
SmolLM3-3B & 3B & 27.78$\mid$11.55 & 30.00$\mid$13.36 & 35.26$\mid$14.20 & 83.48$\mid$79.15 & \textbf{71.21}$\mid$03.55 & 90.80$\mid$80.20 & & 56.42$\mid$33.67 \\
Phi-4-mini-reasoning & 4B & 30.30$\mid$14.55 & 40.00$\mid$15.41 & 32.37$\mid$18.39 & 87.10$\mid$85.54 & 51.58$\mid$22.05 & 90.80$\mid$79.84 & & 55.36$\mid$39.30 \\
Qwen3-4B-Thinking-2507 & 4B & 67.17$\mid$28.48 & 73.33$\mid$05.93 & 04.62$\mid$00.23 & 89.05$\mid$81.77 & 38.57$\mid$20.79 & 97.60$\mid$57.08 & & 61.72$\mid$32.38 \\
\midrule
\textbf{Frugal-Thinking-30B-A3B-Stage-1 (ours)} & 30B &
\makecell{70.20$\mid$39.14\\{\scriptsize \textcolor[HTML]{D00000}{(-0.51)}$\mid$\textcolor[HTML]{D00000}{(-04.82)}}} &
\makecell{83.33$\mid$15.41\\{\scriptsize \textcolor[HTML]{D00000}{(-03.33)}$\mid$\textcolor[HTML]{009937}{(+01.48)}} }&
\makecell{06.94$\mid$00.72\\{\scriptsize \textcolor[HTML]{D00000}{(-01.15)}$\mid$\textcolor[HTML]{009937}{(+00.09)}}} &
\makecell{90.47$\mid$87.79\\{\scriptsize \textcolor[HTML]{009937}{(+00.18)}$\mid$\textcolor[HTML]{D00000}{(-02.50)}}} &
\makecell{41.65$\mid$40.54\\{\scriptsize \textcolor[HTML]{009937}{(+00.30)}$\mid$\textcolor[HTML]{D00000}{(-00.81)}}} &
\makecell{97.20$\mid$73.26\\{\scriptsize \textcolor[HTML]{D00000}{(-00.60)}$\mid$\textcolor[HTML]{009937}{(+10.53)}}} &
& \makecell{64.97$\mid$42.8\\{\scriptsize \textcolor[HTML]{D00000}{(-00.85)}$\mid$\textcolor[HTML]{009937}{(+00.66)}}} \\

\textbf{Frugal-Thinking-30B-A3B-Stage-2 (ours)} & 30B &
\makecell{65.65$\mid$33.17\\{\scriptsize \textcolor[HTML]{D00000}{(-05.06)}$\mid$\textcolor[HTML]{D00000}{(-10.79)}}} &
\makecell{\textbf{86.67}$\mid$44.60\\{\scriptsize \textcolor[HTML]{009937}{(+00.00)}$\mid$\textcolor[HTML]{009937}{(+30.67)}}} &
\makecell{46.24$\mid$21.62\\{\scriptsize \textcolor[HTML]{009937}{(+38.15)}$\mid$\textcolor[HTML]{009937}{(+20.99)}}} &
\makecell{\textbf{90.57}$\mid$75.55\\{\scriptsize \textcolor[HTML]{D00000}{(-00.28)}$\mid$\textcolor[HTML]{D00000}{(-15.14)}}}&
\makecell{42.07$\mid$36.92\\{\scriptsize \textcolor[HTML]{009937}{(+00.72)}$\mid$\textcolor[HTML]{D00000}{(-04.43)}}} &
\makecell{97.40$\mid$88.78\\{\scriptsize \textcolor[HTML]{D00000}{(-00.40)}$\mid$\textcolor[HTML]{009937}{(+26.05)}}} &
& \makecell{\textbf{71.43}$\mid$50.11\\{\scriptsize \textcolor[HTML]{009937}{(+05.61)}$\mid$\textcolor[HTML]{009937}{(+07.96)}}} \\

\textbf{Frugal-Thinking-4B-Stage-1 (ours)} & 4B &
\makecell{63.64$\mid$42.21\\{\scriptsize \textcolor[HTML]{D00000}{(-03.53)}$\mid$\textcolor[HTML]{009937}{(+13.73)}}} &
\makecell{60.00$\mid$46.02\\{\scriptsize \textcolor[HTML]{D00000}{(-13.33)}$\mid$\textcolor[HTML]{009937}{(+40.27)}}} &
\makecell{35.84$\mid$31.54\\{\scriptsize \textcolor[HTML]{009937}{(+31.22)}$\mid$\textcolor[HTML]{009937}{(+31.31)}}} &
\makecell{89.24$\mid$76.59\\{\scriptsize \textcolor[HTML]{009937}{(+00.19)}$\mid$\textcolor[HTML]{D00000}{(-04.58)}}} &
\makecell{39.91$\mid$22.43\\{\scriptsize \textcolor[HTML]{009937}{(+01.34)}$\mid$\textcolor[HTML]{009937}{(+01.64)}}} &
\makecell{95.00$\mid$86.30\\{\scriptsize \textcolor[HTML]{D00000}{(-02.60)}$\mid$\textcolor[HTML]{009937}{(+29.22)}} }&
& \makecell{63.94$\mid$50.85\\{\scriptsize \textcolor[HTML]{009937}{(+02.22)}$\mid$\textcolor[HTML]{009937}{(+18.47)}} }\\

\textbf{Frugal-Thinking-4B-Stage-2 (ours)} & 4B &
\makecell{70.20$\mid$53.84\\{\scriptsize \textcolor[HTML]{009937}{(+03.03)}$\mid$\textcolor[HTML]{009937}{(+25.36)}}} &
\makecell{70.00$\mid$\textbf{70.00}\\{\scriptsize \textcolor[HTML]{D00000}{(-03.33)}$\mid$\textcolor[HTML]{009937}{(+64.07)}}} &
\makecell{47.40$\mid$\textbf{47.40}\\{\scriptsize \textcolor[HTML]{009937}{(+42.78)}$\mid$\textcolor[HTML]{009937}{(+47.17)}}} &
\makecell{89.00$\mid$80.06\\ {\scriptsize \textcolor[HTML]{D00000}{(-00.05)}$\mid$\textcolor[HTML]{D00000}{(-01.11)}}}
 &
\makecell{39.49$\mid$23.20\\{\scriptsize \textcolor[HTML]{009937}{(+00.92)}$\mid$\textcolor[HTML]{009937}{(+02.41)}}} &
\makecell{95.20$\mid$\textbf{95.20}\\{\scriptsize \textcolor[HTML]{D00000}{(-02.40)}$\mid$\textcolor[HTML]{009937}{(+33.12)}}} &
& \makecell{68.55$\mid$\textbf{61.22}\\{\scriptsize \textcolor[HTML]{009937}{(+06.83)}$\mid$\textcolor[HTML]{009937}{(+28.84)}}} \\
\bottomrule
\end{tabular}%
}
\caption{Reasoning benchmark performance under a 42k-token decoding budget. For each cell, the left value reports pass@1 (except on IFEval, where it reports average accuracy), and the right value reports $EAA_{3}$. For our Frugal-Thinking models, two additional values appear beneath the accuracy, indicating the absolute difference relative to the corresponding base model.}
\label{tab:reasoning_benchmarks}
\end{table*}

\begin{table*}[ht]
\centering
\renewcommand{\arraystretch}{1.2} 
\resizebox{\textwidth}{!}{%
\begin{tabular}{l ccccccc c c}
\toprule
\textbf{Model} & 
\textbf{Size} &
\textbf{GPQA Diamond} &
\textbf{AIME25} &
\textbf{Omni-Hard} &
\textbf{GSM\_PLUS} &
\textbf{IFEVAL} &
\textbf{MATH\_500} &
\phantom{a} &
\textbf{Average Length} \\
\midrule
Qwen3-30B-A3B-Thinking-2507  & 30B & 7208.61 & 17887.8 & 26960.1 & \textbf{1373.03} & \textbf{1179.44} & 5069.94 & & 9946.49 \\
Magistral-Small-2509  & 24B & \textbf{5130.58} & 15666.5 & 20864.1 & 1509.53 & 1548.66 & 3878.34 & & 8099.62 \\
Magistral-Small-2507  & 24B & 18247 & 23349 & 25133.8 & 16104.4 & 3815.28 & 16046.7 & & 17116.03 \\
SmolLM3-3B                  & 3B  & 8966.65 & 13136.2 & 17076.9 & 1634.25 & 5521.41 & 3695.5 & & 8338.48 \\
Phi-4-mini-reasoning        & 4B  & 8338.75 & 13811.7 & 15009.4 & 1461.65 & 2409.01 & 3714.96 & & 7457.58 \\
Qwen3-4B-Thinking-2507       & 4B  & 8882.41 & 21090.1 & 29642.2 & 1791.69 & 2073.93 & 5465.89 & & 11491.04 \\
\midrule
\textbf{Frugal-Thinking-30B-A3B-Stage-1 (ours)} & 30B  & 7684.76 & 17232.7 & 25188.2 & 1521.09 & 1218.65 & 4376.94 & & 9537.06 \\
\textbf{Frugal-Thinking-30B-A3B-Stage-2 (ours)} & 30B  & 8115.41 & 12464 & 16181.6 & 2263.64 & 1368.3 & 3560.28 & & 7325.54 \\
\textbf{Frugal-Thinking-4B-Stage-1 (ours)} & 4B  & 6925.65 & 10604.1 & 12380.3 & 2123.68 & 2013.00 & 3574.92 & & 6270.28 \\
\textbf{Frugal-Thinking-4B-Stage-2 (ours)} & 4B  & 6290.44 & \textbf{9367.67} & \textbf{11611.9} & 1892.89 & 1949.61 & \textbf{3162.40} & & \textbf{5712.49} \\
\bottomrule
\end{tabular}%
}
\caption{Average output length (tokens) per benchmark under the 42k-token decoding budget.}
\label{tab:reasoning_lengths}
\end{table*}

\subsection{Reasoning Benchmarks}

We evaluate models on diverse reasoning benchmarks spanning mathematics, STEM, and instruction following.

\smallskip

\noindent\textit{\underline{Mathematics}}
        \begin{itemize}
            \item \textbf{AIME25} \citep{ye2025aime}: The 2025 American Invitational Mathematics Examination, containing 30 integer-answer problems.
            \item \textbf{Omni-MATH-Hard} \citep{gao2024omnimathuniversalolympiadlevel}: The hardest subset of Omni-MATH, retaining only Olympiad-level problems rated 9–10 in difficulty (100 problems total).
            \item \textbf{MATH-500}\footnote{\url{https://hf.co/datasets/HuggingFaceH4/MATH-500}}: A held-out set of 500 problems from the original MATH benchmark introduced in “Let’s Verify Step by Step” \citep{lightman2023letsverifystepstep}.
            \item \textbf{GSM-Plus} \citep{li2024gsmpluscomprehensivebenchmarkevaluating}: A robustness extension of GSM8K with controlled perturbations (e.g., rewording, distractors, numerical changes) to assess consistency under input variations.
        \end{itemize}
        
\noindent\textit{\underline{STEM}}
        \begin{itemize}
            \item \textbf{GPQA-Diamond} \citep{rein2023gpqagraduatelevelgoogleproofqa}: 198 expert-written, “Google-proof” multiple-choice questions across biology, physics, and chemistry.
        \end{itemize}

\noindent\textit{\underline{Instruction Following}}
        \begin{itemize}
            \item \textbf{IFEval} \citep{zhou2023instructionfollowingevaluationlargelanguage}: 500 prompts designed to test precise adherence to explicit textual instructions with verifiable outcomes.
        \end{itemize}

\subsection{Metrics}



\begin{definition}[Efficiency Adjusted Accuracy (EAA)]
\label{def:eaa}
To jointly evaluate reasoning accuracy and conciseness, we define the \emph{Efficiency Adjusted Accuracy} (EAA) metric.
Let $a \in [0,1]$ denote the pass@k (or accuracy) and $L \in [L_{\min}, L_{\max}]$ the mean output length in tokens of a model on a given benchmark. 
For a tunable penalty exponent $\gamma > 0$, we define
\begin{equation}
\mathrm{EAA}_\gamma(a,L) = a \cdot \exp \left[ - \gamma \cdot \left(\frac{L - L_{\min}}{L_{\max} - L_{\min}}\right)\right].
\label{eq:eaa}
\end{equation}
This formulation linearly rescales output length to the unit interval, so that shorter completions ($L \!\approx\! L_{\min}$) preserve the full score, whereas longer ones ($L \!\approx\! L_{\max}$) are increasingly penalized depending on $\gamma$.
\end{definition}

\smallskip

\noindent\textit{\underline{Interpretation}.}
$\mathrm{EAA}_\gamma$ remains bounded in $[0,1]$ and decreases monotonically with $L$, with larger $\gamma$ penalizing verbosity more strongly.
\subsection{Results}

We refer to the 4B models as \texttt{Frugal-Thinking- 4B-Stage-1} and \texttt{Frugal-Thinking-4B-Stage-2}, and the 30B-A3B models as \texttt{Frugal- Thinking-30B-A3B-Stage-1} and \texttt{Frugal- Thinking-30B-A3B-Stage-2}, for Stage 1 and Stage 2 respectively.

Table \ref{tab:reasoning_benchmarks} summarizes the reasoning performance of models ranging from 3B to 30B parameters under a 42k-token decoding limit, while Table \ref{tab:reasoning_lengths} reports their corresponding average output lengths. Each cell in Table \ref{tab:reasoning_benchmarks} contains two metrics: the left value is standard Pass@1 (accuracy using LLM-as-a-Judge for Omni-Hard and average scores for IFEval), while the right value is EAA (accuracy normalized by output length) using $\gamma$=3.0.

At this maximum context length, \texttt{Frugal-Thinking-4B-Stage-2} achieves an average accuracy of 68.55\% and an EAA of 61.22, outperforming its base model (61.72 / 32.38) by +6.83 and +28.84, respectively. While \texttt{Frugal-Thinking-30B-A3B-Stage-2} demonstrates the same behavior achieving 71.43 / 50.11 on average compared to its base model (65.82 / 42.15). The Stage 1 variant of our 4B model also improves to 63.94 / 50.85, showing that our Stage 1 fine-tuning yields substantially better token efficiency. However, Stage 1 was less effective on the 30B-A3B variant with regard to token efficiency. Compared to larger or similar-sized baselines, the 24B \texttt{Magistral} models achieve 70.16 / 48.95 and 60.12 / 04.09 for their 2509 and 2507 versions respectively. While \texttt{Phi-4-mini-reasoning} and \texttt{SmolLM3-3B} trail behind at 55.36 / 39.30 and 56.42 / 33.67, confirming that the \texttt{Frugal-Thinking} models preserve or slightly improve overall accuracy while delivering far better efficiency.

The length analysis in Table \ref{tab:reasoning_lengths} reinforces this efficiency narrative. While \texttt{Qwen3-4B-Thinking- 2507} generates on average 11491 tokens per sample, our \texttt{Frugal-Thinking-4B} variants drastically reduce this to 6270 for Stage 1 and 5712 for Stage 2.  
The efficiency gains are most pronounced on harder mathematical reasoning tasks, notably AIME25 and Omni-Hard, where solution chains are typically long. On these benchmarks, Stage 2 of our 4B variant achieves comparable or higher accuracy using 55–61\% fewer tokens, indicating that it learns to reason more efficiently, while still reaching correct final answers. In contrast, for easier arithmetic problems such as GSM\_PLUS, where all models already reach high accuracy with short outputs, the advantage is less pronounced; Stage 2’s generations are slightly longer (+5.6\%) and EAA shows a small regression. This pattern suggests that the \texttt{Frugal-Thinking} models allocate reasoning effort adaptively—compressing complex reasoning when needed but not over-optimizing brevity on tasks that are inherently simple.

Table \ref{tab:avg-accuracy-by-context} illustrates how different models behave under increasing generation budgets (8k → 16k → 32k → 42k) For AIME25.
Our \texttt{Frugal-Thinking-4B} models, particularly Stage 2, demonstrate superior efficiency at lower budgets. At 8k and 16k tokens, they already achieve accuracy levels close to or exceeding larger models.
This indicates that the Frugal models can solve complex, multi-step mathematical problems correctly with much shorter reasoning chains. In contrast, \texttt{Qwen3-4B-Thinking-2507}, \texttt{Qwen3-30B-A3B-Thinking-2507} and \texttt{Magistral -Small-2509} continue to improve with larger decoding budgets (32k and 42k) achieving better accuracies, but their EAA remains consistently lower, suggesting that their accuracy gains rely on significantly longer outputs. 
Overall, these scaling results highlight the strength of our approach on reasoning-intensive tasks, where \texttt{Frugal-Thinking-4B} models maintain a more favorable accuracy-per-token ratio and deliver strong performance even under tight output constraints.

\section{Conclusion}

We show that excessive verbosity in step-by-step reasoning models is largely a consequence of RLVR training practices that overemphasize hard, long-chain problems. Retaining and modestly up-weighting moderately easy problems acts as an implicit length regularizer, exposing the model to short, solvable reasoning trajectories and preventing runaway verbosity.

This simple intervention yields emergent brevity without any explicit length penalty. Models trained under this regime match baseline pass@1 accuracy on AIME25 while producing solutions that are nearly twice as short. Our results highlight a practical path toward more efficient reasoning models by shaping output length through data selection rather than additional constraints.

Future work may extend this idea to other domains such as coding or logical reasoning, explore adaptive curricula balancing easy–hard samples, and combine implicit and explicit regularization for finer control of brevity.


\begin{table}[t]
\centering
\renewcommand{\arraystretch}{1.2} 
\resizebox{\columnwidth}{!}{%
\begin{tabular}{lcccc}
\hline
\textbf{Model$\backslash$Context length} & \textbf{8k} & \textbf{16k} & \textbf{32k} & \textbf{42k} \\
\hline
Qwen3-30B-A3B-Thinking-2507               & 20.00 & 60.00 & \textbf{86.67} & \textbf{86.67} \\
Magistral-Small-2509                 & 26.67 & 66.67 & 80.00 & 80.00 \\
Magistral-Small-2507                 & 06.67 & 33.33 & 53.33 & 53.33 \\
SmolLM3-3B                       & 23.33 & 30.00 & 30.00 & 30.00 \\
Phi-4-mini-reasoning                 & 23.33 & 33.33 & 40.00 & 40.00 \\
Qwen3-4B-Thinking-2507                    & 13.33 & 46.67 & 73.33 & 73.33 \\
\midrule
Frugal-Thinking-30B-A3B-Stage-1    & 23.33 & 56.67 & \underline{83.33} & \underline{83.33} \\
Frugal-Thinking-30B-A3B-Stage-2 & \underline{40.00} & \textbf{73.33} & \textbf{86.67} & \textbf{86.67} \\
Frugal-Thinking-4B-Stage-1             & 30.00 & 60.00 & 60.00 & 60.00 \\
Frugal-Thinking-4B-Stage-2             & \textbf{53.33} & \underline{70.00} & 70.00 & 70.00 \\
\hline
\end{tabular}}
\caption{AIME25 accuracy by context length.}
\label{tab:avg-accuracy-by-context}
\end{table}


\section*{Limitations}

Our study focuses on math reasoning tasks with verifiable rewards, evaluated on both dense and MoE models (4B and 30B respectively).
While we provide information-theoretic arguments motivating reduced verbosity, the emergence of brevity is primarily supported by empirical evidence rather than a complete theoretical characterization.
A deeper theoretical understanding of this behavior remains an important direction for future work.

\bibliography{references}

\begin{thebibliography}{13}
\providecommand{\natexlab}[1]{#1}

\bibitem[{Gao et~al.(2024)Gao, Song, Yang, Cai, Miao, Dong, Li, Ma, Chen, Xu,
  Tang, Wang, Zan, Quan, Zhang, Sha, Zhang, Ren, Liu, and
  Chang}]{gao2024omnimathuniversalolympiadlevel}
Bofei Gao, Feifan Song, Zhe Yang, Zefan Cai, Yibo Miao, Qingxiu Dong, Lei Li,
  Chenghao Ma, Liang Chen, Runxin Xu, Zhengyang Tang, Benyou Wang, Daoguang
  Zan, Shanghaoran Quan, Ge~Zhang, Lei Sha, Yichang Zhang, Xuancheng Ren,
  Tianyu Liu, and Baobao Chang. 2024.
\newblock \href {https://arxiv.org/abs/2410.07985} {Omni-math: A universal
  olympiad level mathematic benchmark for large language models}.
\newblock \emph{Preprint}, arXiv:2410.07985.

\bibitem[{He et~al.(2025)He, Liang, Xu, Liu, Chen, Wang, Song, Yu, Liang, Wang
  et~al.}]{deepmath}
Zhiwei He, Tian Liang, Jiahao Xu, Qiuzhi Liu, Xingyu Chen, Yue Wang, Linfeng
  Song, Dian Yu, Zhenwen Liang, Wenxuan Wang, and 1 others. 2025.
\newblock Deepmath-103k: A large-scale, challenging, decontaminated, and
  verifiable mathematical dataset for advancing reasoning.
\newblock \emph{arXiv preprint arXiv:2504.11456}.

\bibitem[{Ji et~al.(2025{\natexlab{a}})Ji, Tian, Zhao, Wang, Chen, Peng, Zhao,
  and Li}]{amthinkingv1_advancingfrontierreasoning}
Yunjie Ji, Xiaoyu Tian, Sitong Zhao, Haotian Wang, Shuaiting Chen, Yiping Peng,
  Han Zhao, and Xiangang Li. 2025{\natexlab{a}}.
\newblock \href {https://arxiv.org/abs/2505.08311} {Am-thinking-v1: Advancing
  the frontier of reasoning at 32b scale}.
\newblock \emph{Preprint}, arXiv:2505.08311.

\bibitem[{Ji et~al.(2025{\natexlab{b}})Ji, Zhao, Tian, Wang, Chen, Peng, Zhao,
  and Li}]{am_team_difficultyawarestagedreinforcementlearning}
Yunjie Ji, Sitong Zhao, Xiaoyu Tian, Haotian Wang, Shuaiting Chen, Yiping Peng,
  Han Zhao, and Xiangang Li. 2025{\natexlab{b}}.
\newblock \href {https://arxiv.org/abs/2504.00829} {How difficulty-aware staged
  reinforcement learning enhances llms' reasoning capabilities: A preliminary
  experimental study}.
\newblock \emph{Preprint}, arXiv:2504.00829.

\bibitem[{Li et~al.(2024)Li, Cui, Zhao, Kong, and
  Bi}]{li2024gsmpluscomprehensivebenchmarkevaluating}
Qintong Li, Leyang Cui, Xueliang Zhao, Lingpeng Kong, and Wei Bi. 2024.
\newblock \href {https://arxiv.org/abs/2402.19255} {Gsm-plus: A comprehensive
  benchmark for evaluating the robustness of llms as mathematical problem
  solvers}.
\newblock \emph{Preprint}, arXiv:2402.19255.

\bibitem[{Lightman et~al.(2023)Lightman, Kosaraju, Burda, Edwards, Baker, Lee,
  Leike, Schulman, Sutskever, and Cobbe}]{lightman2023letsverifystepstep}
Hunter Lightman, Vineet Kosaraju, Yura Burda, Harri Edwards, Bowen Baker, Teddy
  Lee, Jan Leike, John Schulman, Ilya Sutskever, and Karl Cobbe. 2023.
\newblock \href {https://arxiv.org/abs/2305.20050} {Let's verify step by step}.
\newblock \emph{Preprint}, arXiv:2305.20050.

\bibitem[{Mistral-AI et~al.(2025)Mistral-AI, :, Rastogi, Jiang, Lo, Berrada,
  Lample, Rute, Barmentlo, Yadav, Khandelwal, Chandu, Blier, Saulnier, Dinot,
  Darrin, Gupta, Soletskyi, Vaze, Scao, Wang, Yang, Liu, Sablayrolles, Héliou,
  Martin, Ehrenberg, Agarwal, Roux, Darcet, Mensch, Bout, Rozière, Monicault,
  Bamford, Wallenwein, Renaudin, Lanfranchi, Dabert, Mizelle, de~las Casas,
  Chane-Sane, Fugier, Hanna, Delerce, Guinet, Novikov, Martin, Jaju,
  Ludziejewski, Chabran, Delignon, Studnia, Amar, Roberts, Denize, Saxena,
  Jain, Zhao, Martin, Gao, Lavaud, Pellat, Guillaumin, Felardos, Augustin,
  Seznec, Raghuraman, Duchenne, Wang, von Platen, Saffer, Jacob, Wambergue,
  Kurylowicz, Muddireddy, Chagniot, Stock, Agrawal, Sauvestre, Delacourt,
  Gandhi, Subramanian, Dalal, Gandhi, Ghosh, Mishra, Aithal, Antoniak,
  Schueller, Lavril, Robert, Wang, Lacroix, Nemychnikova, Paltz, Richard, Li,
  Marshall, Zhang, and Tang}]{magistral}
Mistral-AI, :, Abhinav Rastogi, Albert~Q. Jiang, Andy Lo, Gabrielle Berrada,
  Guillaume Lample, Jason Rute, Joep Barmentlo, Karmesh Yadav, Kartik
  Khandelwal, Khyathi~Raghavi Chandu, Léonard Blier, Lucile Saulnier, Matthieu
  Dinot, Maxime Darrin, Neha Gupta, Roman Soletskyi, Sagar Vaze, and 82 others.
  2025.
\newblock \href {https://arxiv.org/abs/2506.10910} {Magistral}.
\newblock \emph{Preprint}, arXiv:2506.10910.

\bibitem[{Rein et~al.(2023)Rein, Hou, Stickland, Petty, Pang, Dirani, Michael,
  and Bowman}]{rein2023gpqagraduatelevelgoogleproofqa}
David Rein, Betty~Li Hou, Asa~Cooper Stickland, Jackson Petty, Richard~Yuanzhe
  Pang, Julien Dirani, Julian Michael, and Samuel~R. Bowman. 2023.
\newblock \href {https://arxiv.org/abs/2311.12022} {Gpqa: A graduate-level
  google-proof q\&a benchmark}.
\newblock \emph{Preprint}, arXiv:2311.12022.

\bibitem[{Schulman et~al.(2017)Schulman, Wolski, Dhariwal, Radford, and
  Klimov}]{ppo}
John Schulman, Filip Wolski, Prafulla Dhariwal, Alec Radford, and Oleg Klimov.
  2017.
\newblock \href {https://arxiv.org/abs/1707.06347} {Proximal policy
  optimization algorithms}.
\newblock \emph{Preprint}, arXiv:1707.06347.

\bibitem[{Shao et~al.(2024)Shao, Wang, Zhu, Xu, Song, Bi, Zhang, Zhang, Li, Wu,
  and Guo}]{grpo}
Zhihong Shao, Peiyi Wang, Qihao Zhu, Runxin Xu, Junxiao Song, Xiao Bi, Haowei
  Zhang, Mingchuan Zhang, Y.~K. Li, Y.~Wu, and Daya Guo. 2024.
\newblock \href {https://arxiv.org/abs/2402.03300} {Deepseekmath: Pushing the
  limits of mathematical reasoning in open language models}.
\newblock \emph{Preprint}, arXiv:2402.03300.

\bibitem[{Sheng et~al.(2025)Sheng, Zhang, Ye, Wu, Zhang, Zhang, Peng, Lin, and
  Wu}]{verl}
Guangming Sheng, Chi Zhang, Zilingfeng Ye, Xibin Wu, Wang Zhang, Ru~Zhang,
  Yanghua Peng, Haibin Lin, and Chuan Wu. 2025.
\newblock Hybridflow: A flexible and efficient rlhf framework.
\newblock In \emph{Proceedings of the Twentieth European Conference on Computer
  Systems}, pages 1279--1297.

\bibitem[{Ye et~al.(2025)Ye, Xiao, Mi, and Liu}]{ye2025aime}
Yixin Ye, Yang Xiao, Tiantian Mi, and Pengfei Liu. 2025.
\newblock Aime-preview: A rigorous and immediate evaluation framework for
  advanced mathematical reasoning.

\bibitem[{Zhou et~al.(2023)Zhou, Lu, Mishra, Brahma, Basu, Luan, Zhou, and
  Hou}]{zhou2023instructionfollowingevaluationlargelanguage}
Jeffrey Zhou, Tianjian Lu, Swaroop Mishra, Siddhartha Brahma, Sujoy Basu,
  Yi~Luan, Denny Zhou, and Le~Hou. 2023.
\newblock \href {https://arxiv.org/abs/2311.07911} {Instruction-following
  evaluation for large language models}.
\newblock \emph{Preprint}, arXiv:2311.07911.

\end{thebibliography}

\newpage
\appendix
\onecolumn

\end{document}